\documentclass[letterpaper]{article} %
\usepackage[preprint]{aaai2027}  %
\usepackage[hyphens]{url}  %
\usepackage{graphicx} %
\usepackage{natbib}  %
\usepackage{caption} %
\usepackage{algorithm}
\usepackage{algorithmic}
\usepackage{xspace}
\usepackage{amsmath}
\usepackage{newfloat}
\usepackage{listings}
\usepackage{pifont}
\usepackage{amssymb}
\usepackage{multirow}
\usepackage[table]{xcolor}
\DeclareCaptionStyle{ruled}{labelfont=normalfont,labelsep=colon,strut=off} %
\floatstyle{ruled}
\newfloat{listing}{tb}{lst}{}
\floatname{listing}{Listing}

\usepackage{booktabs}
\usepackage{tabularx}
\usepackage{makecell}
\usepackage[most,skins]{tcolorbox}
\tcbuselibrary{listings, breakable}

\definecolor{aaibest}{HTML}{A2D8A2}        %
\definecolor{aaisecond}{HTML}{FCEAB4}      %
\newcommand{\best}[1]{\cellcolor{aaibest}\textbf{#1}}
\newcommand{\second}[1]{\cellcolor{aaisecond}\textbf{#1}}

\title{When Metrics Reward the Worst Translations: Internalizing Cultural Reasoning for Social Media Translation Evaluation}
\author{
    Yiwen Qiu\textsuperscript{\rm 1},
    Linjuan Wu\textsuperscript{\rm 1},
    Dingming Li\textsuperscript{\rm 1},
    Yizhou Liu\textsuperscript{\rm 1},
    Zixuan Wang\textsuperscript{\rm 1},
    Haolei Xu\textsuperscript{\rm 1},
    Ye Guo\textsuperscript{\rm 2},
    Daoxin Zhang\textsuperscript{\rm 2},
    Weiming Lu\textsuperscript{\rm 1},
    Yongliang Shen\textsuperscript{\rm 1}\textsuperscript{\dag}
}
\affiliations{
    \textsuperscript{\rm 1}Zhejiang University\\
    \textsuperscript{\rm 2}Xiaohongshu Inc.
}

\begin{document}
\captionsetup{labelfont=normalfont,textfont=normalfont}

\maketitle

\begin{abstract}
    Automatic translation quality metrics trained on general-domain corpora systematically fail on social media content, where communicative intent is encoded in culturally loaded expressions (internet slang, homophonic ciphers, and platform-specific idioms) rather than surface token patterns. We conduct a systematic empirical analysis demonstrating that standard metrics including COMET, XCOMET, and BERTScore exhibit near-zero or negative correlation with human cultural judgments, and even display a severity inversion in which scores increase as translation quality deteriorates. We further show that this failure extends to large language model judges: Qwen3-235B achieves Cohen's $\kappa$ of only $0.162$, revealing that the bottleneck is not reasoning capacity but cultural grounding: models lack the domain-specific cultural knowledge needed to identify which aspects of a translation require scrutiny. To address this, we propose CuRIL, a reinforcement learning framework that internalizes cultural reasoning: cultural annotations are prepended inside the model's reasoning, excluded from policy gradients via a token-level loss mask, and injected with a probability that decays to zero over training, progressively forcing autonomous cultural judgment. On a 1,444-sample human-annotated social media translation benchmark, Qwen3-8B trained with CuRIL achieves Cohen's $\kappa = 0.370$ and Exact Match accuracy of 45.22\%, approaching Gemini-3.1-Pro with $30\times$ fewer parameters and surpassing models up to 235B in scale. We further demonstrate that our judge produces reliable reward signals for downstream translation optimization, reducing the low-quality translation rate by over 20 percentage points under independent evaluation.
\end{abstract}

\section{Introduction}
\label{sec:intro}

\begin{figure}[t]
    \centering
    \includegraphics[width=\columnwidth]{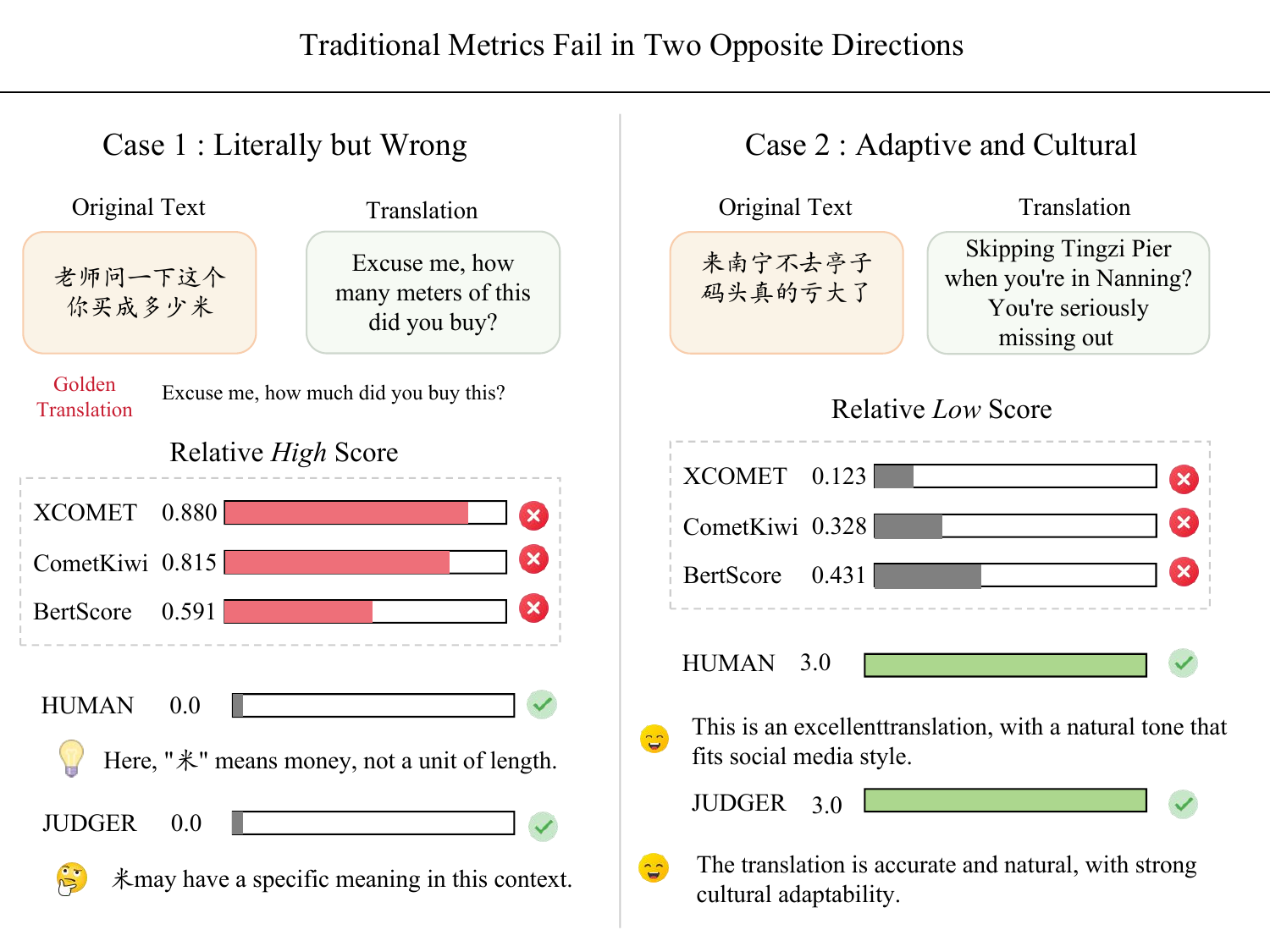}
     \caption{Two representative cases on social media translation: a literal but culturally wrong translation (left) and a culturally adaptive translation (right), scored by automatic metrics, our judge, and human annotators.}
    \label{fig:motivation}
\end{figure}

Machine translation quality has advanced rapidly~\cite{tian2026beyond,wu2026beyond,yuan2026culture}, yet progress hinges on our ability to \emph{measure} it: automatic metrics and LLM judges supply the reward signals for training and the benchmarks for comparison. When miscalibrated, they silently misdirect optimization.
This risk is acute on social media, where communicative intent is encoded in culturally loaded expressions rather than surface token patterns.
Consider the Chinese social media utterance \textit{l\v{a}osh\={\i} zh\`{e}ge n\v{i} m\v{a}i ch\'{e}ng du\=osh\v{a}o m\v{i}}, literally translated as ``Teacher, how many \textit{meters} of this did you buy?''
As Figure~\ref{fig:motivation} shows, this literal rendering scores 0.880 on XCOMET and 0.591 on BERTScore, yet it is semantically wrong: \textit{m\v{i}} (``meter'') is Chinese internet slang for \textit{qi\'{a}n} (``money''), and the utterance actually asks ``how much did this cost?''
This is not an isolated case but a systematic deficiency.

The root cause is conceptual: social media translation is a \emph{cultural reasoning} problem, not a semantic matching problem~\cite{macko2025multisocial,huang2025can,mei2024slang}.
Metrics such as COMET, XCOMET, and BERTScore, trained on general-domain corpora and optimized for surface similarity, systematically reward literally correct but culturally void translations while penalizing culturally apt paraphrases.
Our empirical study (Section~\ref{sec:preliminary}) on 1,444 human-annotated Chinese--English pairs reveals three compounding failure modes: near-zero correlation with human judgment (Cohen's $\kappa \leq 0.071$), systematic invisibility of cultural errors, and a \emph{severity inversion} in which metric scores \textit{increase} as translation quality deteriorates.

Turning to LLM-as-a-Judge~\cite{kocmi2023gemba,zhan2026large} does not resolve this failure.
Such judges are increasingly used as reward signals for translation training~\cite{feng2025mt,wang2026deeptrans,feng2025mt-zero}, which raises the stakes of their miscalibration; yet even Qwen3-235B achieves Cohen's $\kappa$ of only $0.162$ on our benchmark.
The bottleneck is not reasoning \emph{capacity} but \emph{cultural grounding}: models lack the domain-specific cultural knowledge needed to recognize culturally loaded expressions (internet slang, homophonic ciphers, platform-specific idioms) and therefore cannot assess whether a translation preserves communicative intent~\cite{moghe2025machine,rikters2024entity,zhou2026incentivizing}.
A diagnostic experiment confirms this: injecting brief cultural annotations (\emph{translation hints}) at test time, without fine-tuning, boosts Qwen3-8B from EM $31.60\%$ to $44.35\%$, demonstrating that models can perform accurate cultural evaluation once the relevant context is supplied.

However, providing per-sample hints at inference time requires expert annotation or a retrieval pipeline for every query, which is infeasible in production.
This motivates our central question: \emph{can cultural reasoning be internalized into model parameters, eliminating the dependency on external hints?}

We propose \textbf{CuRIL} (\textbf{Cu}ltural \textbf{R}easoning \textbf{I}nternalization via curriculum \textbf{L}earning), a reinforcement learning framework built on GRPO with two mechanisms: (1)~translation hints injected as a \textbf{masked prefix} in the model's reasoning block, excluded from policy gradient computation; and (2)~a \textbf{linear decay schedule} that progressively removes hints, forcing the model toward autonomous cultural judgment.
On Qwen3-8B, CuRIL achieves $\kappa = 0.370$ and EM $= 45.22\%$, approaching Gemini~3.1~Pro~\cite{google-gemini-3} ($\kappa = 0.438$) with $30\times$ fewer parameters, and surpassing GPT-5.5~\cite{gpt-5} ($\kappa = 0.338$) and Qwen3-235B~\cite{qwen3} ($\kappa = 0.162$).
A downstream translation model trained with CuRIL reward signals further reduces the low-quality translation rate from 25.6\% to 4.9\% under independent evaluation.

Our contributions are:
\begin{itemize}

\item We establish that social media translation evaluation is a cultural reasoning task, not a semantic matching problem, and provide systematic empirical evidence that mainstream metrics fail due to a structural absence of cultural grounding, not insufficient precision or scale.

\item We propose CuRIL, a general framework for internalizing external reasoning signals into model parameters via reinforcement learning with decaying scaffolds, eliminating the need for expert annotation or retrieval at inference time.

\item CuRIL-trained 8B models match frontier closed-source systems and surpass models up to $30\times$ larger; the resulting judge further reduces the low-quality translation rate from 25.6\% to 4.9\% downstream, establishing a virtuous cycle between better evaluation and better translation.

\end{itemize}

\section{Preliminary Study}
\label{sec:preliminary}

We quantify how existing metrics fail on social media translation, and why, on a human-annotated validation set of 1,444 Chinese--English pairs.
We compare quality-estimation-based CometKiwi and XCOMET and token-similarity-based BERTScore against human judgments, which follow a four-point rubric (0--3) covering semantic accuracy, cultural appropriateness, and format compliance.

\subsection{Metrics Cannot Distinguish Good from Bad}
\label{sec:finding1}

\begin{figure}[t]
    \centering
    \includegraphics[width=\columnwidth]{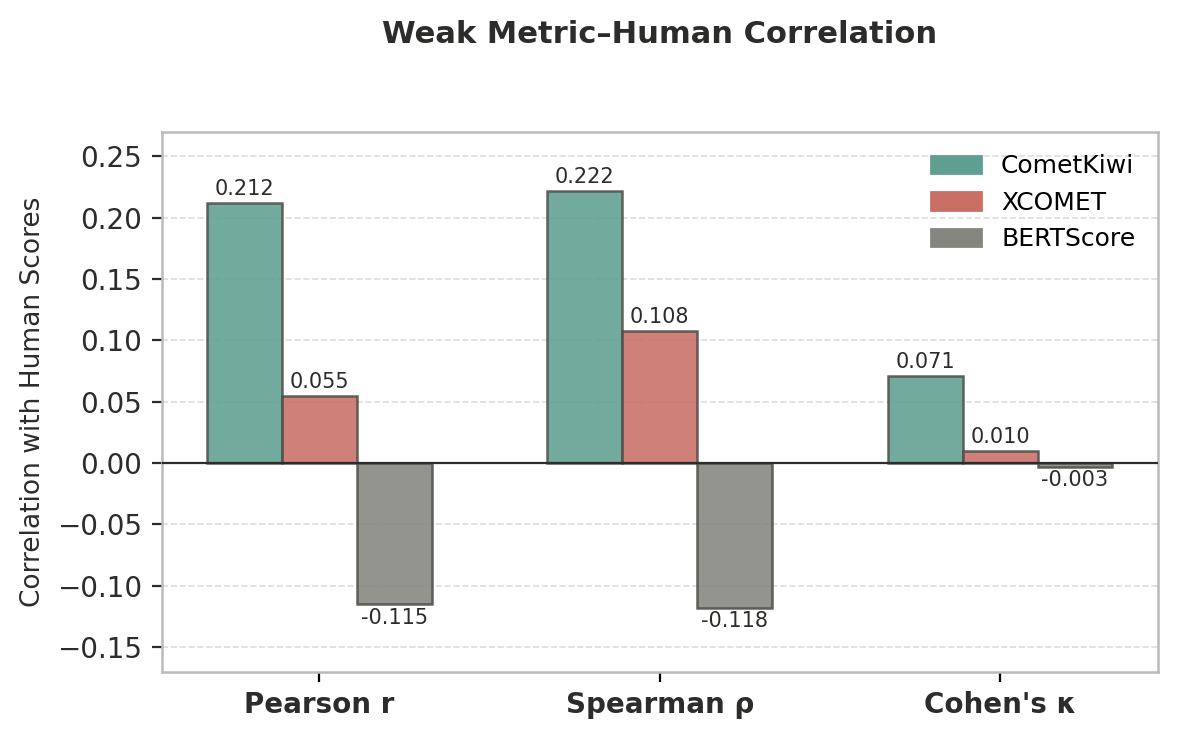}
     \caption{Correlation between automatic metrics and human judgments on the validation set.}
    \label{fig:preliminary_one}
\end{figure}

Figure~\ref{fig:preliminary_one} reports the correlation between each metric and human judgments.
The results are stark: CometKiwi achieves the highest Pearson correlation at $r = 0.212$, XCOMET reaches only $r = 0.055$, and BERTScore records $r = {-}0.115$, a \emph{negative} correlation indicating that higher BERTScore is systematically associated with \emph{lower} human ratings.
All three metrics achieve Cohen's $\kappa \leq 0.071$, a level conventionally interpreted as no more than slight agreement.

\subsection{Cultural Errors Are Invisible to Metrics}
\label{sec:finding2}

The failure is not randomly distributed.
We define a metric's \emph{blind spot} as samples where the metric scores high ($> 0.7$) but human annotators score low ($\leq 1$).
XCOMET's blind spot covers 29.36\% of all validation samples: nearly a third of the entire validation set consists of low-quality translations that XCOMET certifies as high quality.
Culturally-related errors account for 26\% of the full dataset but rise to 46.9\% within this blind spot, a $1.74\times$ enrichment.
The failures are \emph{systematically concentrated on cultural errors}: the metrics retain partial sensitivity to surface errors such as grammatical mistakes but lose discriminative ability precisely where cultural reasoning is required.

\subsection{Metrics Reward the Worst Translations}
\label{sec:finding3}

The failure goes beyond mere insensitivity: metrics actively assign \emph{higher} scores to more severe errors.

\begin{table}[!ht]
\centering
\small
\begin{tabularx}{\linewidth}{@{}lXXX@{}}
\toprule
\textbf{Metric} & \textbf{Error-free} & \textbf{Severe} & \textbf{Change} \\
\midrule
Human score  & \best{2.326} & \best{0.287} & $\downarrow$ 87.7\% \\
\midrule
CometKiwi    & \second{0.532} & \second{0.569} & $\uparrow$ 7.0\%   \\
XCOMET       & \second{0.568} & \second{0.743} & $\uparrow$ 30.8\%  \\
BERTScore    & \second{0.573} & \second{0.622} & $\uparrow$ 8.6\%   \\
\bottomrule
\end{tabularx}
\caption{Metric scores at the two extremes of human-rated error severity. \protect\colorbox{aaibest}{\textbf{Green}}: human reference; \protect\colorbox{aaisecond}{\textbf{yellow}}: traditional metrics exhibiting severity inversion.}
\label{tab:severity}
\end{table}

Table~\ref{tab:severity} stratifies samples by human-rated severity.
Human scores move in the expected direction: from $2.326$ for error-free translations to $0.287$ for severely flawed ones, a decline of 87.7\%.
All three metrics move in precisely the opposite direction: CometKiwi increases by 7.0\%, XCOMET by 30.8\%, BERTScore by 8.6\%.

The root cause is a single mechanism underlying all three findings.
Severe errors in social media translation typically arise from \emph{literal translation}, which preserves maximum token-level overlap with the source.
Token overlap is exactly what these metrics, trained on general-domain corpora for semantic adequacy, are optimized to reward.
The result is a systematic \emph{severity inversion}: the metrics do not merely fail to detect the worst translations but actively certify them as the best.
A metric that rewards the worst translations cannot serve as a reliable optimization signal, motivating a departure from similarity-based evaluation toward a culturally grounded approach.

\section{Method}
\label{sec:method}

\begin{figure*}[t]
    \centering
    \includegraphics[width=\textwidth]{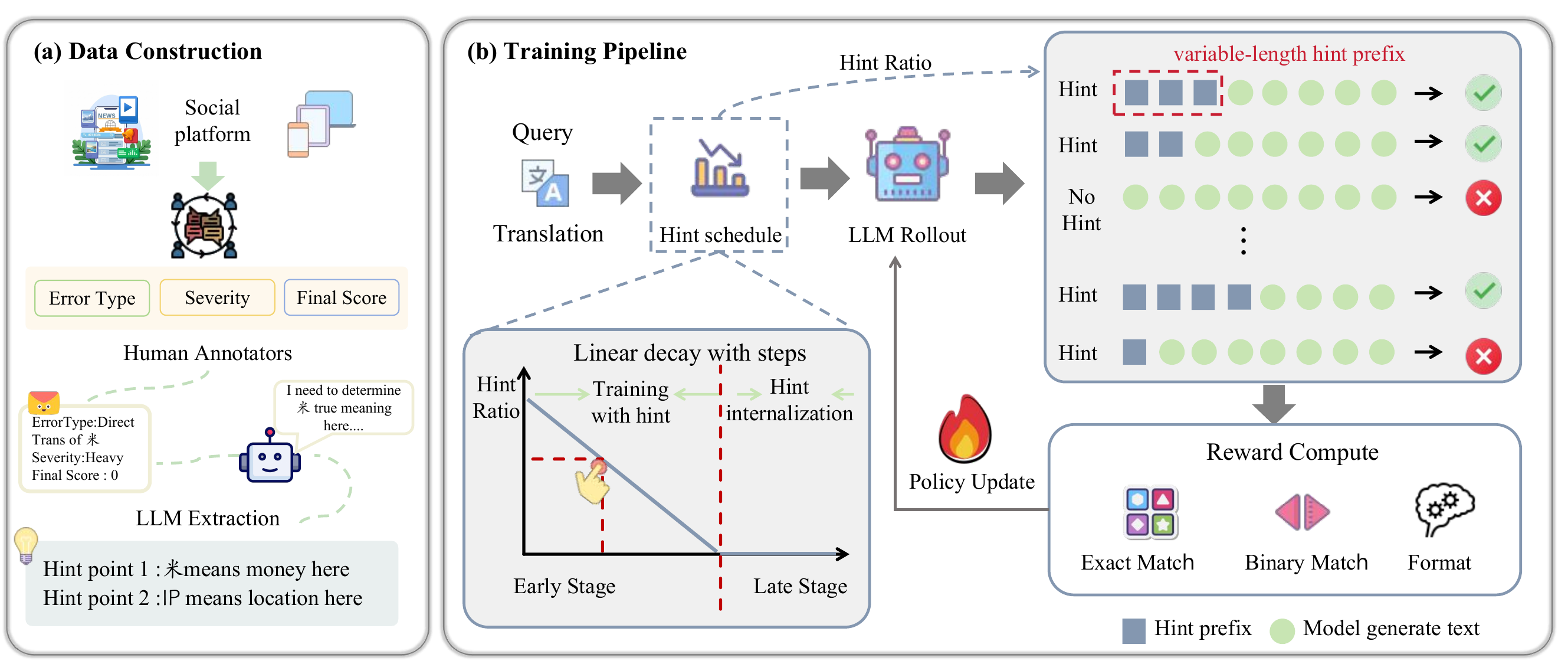}
     \caption{Overview of CuRIL. (a) Data construction: human annotation of error types, severities, and scores, followed by LLM-based hint extraction. (b) Training: each query is augmented with a hint prefix under a linearly decaying injection probability; rollouts are scored by deviation-based, binary, and format rewards for the GRPO update.}
    \label{fig:method}
\end{figure*}

To internalize cultural reasoning without expert input at deployment, \textbf{CuRIL} (\textbf{Cu}ltural \textbf{R}easoning \textbf{I}nternalization via curriculum \textbf{L}earning) trains the judge with reinforcement learning in which cultural hints appear only as a gradient-masked, gradually vanishing prefix of the model's own reasoning. The masked prefix guides exploration without receiving policy-gradient credit, and the linearly decaying injection probability progressively forces the model to reproduce the same cultural reasoning on its own. Figure~\ref{fig:method} provides an overview.
We begin by formalizing the task and hint construction,
then describe hint injection with gradient masking, the linear decay schedule,
the reward design, and the optimization objective.

\subsection{Task Formulation}
\label{sec:task}

Let $\mathcal{D} = \{(x_i, y_i, s_i)\}_{i=1}^{N}$ denote the training set,
where $x_i$ is the Chinese source text, $y_i$ is the English translation,
and $s_i \in \mathcal{S} = \{0, 1, 2, 3\}$ is the human quality score
(0: severe error; 3: high-quality translation).
We define two coarse binary quality categories:
$\mathcal{C}_{\text{low}} = \{0, 1\}$ (low quality, unsuitable for use)
and $\mathcal{C}_{\text{high}} = \{2, 3\}$ (high quality, suitable for use).

Given a source--translation pair $(x, y)$ and an optional set of hints $h$,
the judge model $\pi_\theta$ produces a predicted score $\hat{s} \in \mathcal{S}$.
We define the score deviation $\delta = |\hat{s} - s|$ and the binary match indicator
$\mathbf{1}_{\text{same}}(\hat{s},s) = \mathbf{1}[(\hat{s}\ge 2)\;\text{if}\;(s\ge 2)]$,
which equals 1 if $\hat{s}$ and $s$ fall in the same binary category, and 0 otherwise.

\subsection{Hint Construction}
\label{sec:hint}

Translation hints are short, targeted cultural annotations that identify the specific
linguistic features a judge must attend to when evaluating a given translation pair.
Each hint $h_{i,k}$ takes the form of a declarative statement about a culturally-loaded
element---for example, identifying a homophonic cipher or a platform-specific idiom
and clarifying its intended meaning.

For each training sample $(x_i, y_i, s_i)$, an auxiliary LLM $\mathcal{M}_h$ generates $K$ hints:
\begin{equation}
  \mathbf{h}_i = \mathcal{M}_h(x_i,\, y_i,\, s_i)
  = \{h_{i,1},\ldots,h_{i,K}\}.
\end{equation}
The human annotation trace is provided to ground hint generation in the actual evaluation outcome,
so that each hint describes a feature relevant to the quality judgment.
Hints are stored as auxiliary metadata and used exclusively during training;
they are unavailable at inference time.

\subsection{Hint Injection via Response Prefix}
\label{sec:injection}

A key design choice is \emph{how} to deliver hints during training without creating a
distributional mismatch at inference time.
Rather than modifying the input prompt $q=(x,y)$, we prepend hints directly inside
the model's \texttt{<think>} reasoning block as a first-person key-point recap
that mirrors the model's own chain-of-thought style.
The full response token sequence is:
\begin{equation}
  \mathbf{o}_i = \underbrace{[h_1, \ldots, h_{L_h}]}_{\text{hint prefix}} \;\oplus\;
                 \underbrace{[g_1, \ldots, g_m]}_{\text{generated}},
\end{equation}
where $[h_1, \ldots, h_{L_h}]$ denotes the $L_h$ tokens of the rendered hint prefix, $[g_1, \ldots, g_m]$ the $m$ tokens generated by the model, and $\oplus$ denotes concatenation.

\paragraph{Gradient masking.}
To ensure the policy gradient derives solely from the model's own reasoning,
we apply a token-level binary mask $M_{i,t} = \mathbf{1}[t > L_h]$,
giving the masked log-probability:
\begin{equation}
  \log \tilde{\pi}_\theta(\mathbf{o}_i \mid q)
  = \sum_{t=1}^{L_h+m} M_{i,t} \cdot \log \pi_\theta(o_{i,t} \mid q,\, \mathbf{o}_{i,<t}).
\end{equation}
This prevents the model from receiving gradient credit for tokens it did not generate~\cite{zhang2026stephint,huang2025boosting,huang2025blending}.

\subsection{Linear Decay Hint Schedule}
\label{sec:schedule}

If hints were injected throughout training, the model would rely on them as a permanent
external resource rather than internalizing cultural reasoning.
To force internalization, the hint injection probability decays linearly:
\begin{equation}
  p(s) = \max\!\left(0,\; p_0\!\left(1 - \tfrac{s}{T}\right)\right),
  \label{eq:decay}
\end{equation}
where $p_0 \in (0,1]$ is the initial injection rate and $T$ is the decay horizon.
For each of the $n$ rollouts per sample, $z_{i,j}\sim\mathrm{Bernoulli}(p(s))$ independently, where $z_{i,j}=1$ indicates that rollout $j$ is conditioned on the hint prefix,
so each batch contains both hint-conditioned and hint-free responses.
For $s \ge T$, no hints are injected and CuRIL reduces to standard GRPO.

\subsection{Reward Function}
\label{sec:reward}

The reward function $r(\mathbf{o}, s)$ evaluates each rollout in two stages.
Responses failing the format check (\texttt{<think>}/\texttt{<answer>} tags) receive
$r_{\text{format}} = -\lambda$ with no further scoring.
For valid responses:
\begin{equation}
  r(\mathbf{o}, s) = r_{\text{base}}(\delta) + r_{\text{bin}}(\hat{s}, s).
\end{equation}
The \textbf{base reward} penalizes proportionally to the score deviation:
\begin{equation}
  r_{\text{base}}(\delta) =
  \begin{cases}
    \phantom{-}\alpha_0 & \delta = 0 \quad \text{(exact match)}, \\
    \phantom{-}\alpha_1 & \delta = 1, \\
    -\alpha_2           & \delta = 2, \\
    -\alpha_3           & \delta = 3,
  \end{cases}
\end{equation}
with $\alpha_0 > \alpha_1 \ge 0$ and $\alpha_3 > \alpha_2 \ge 0$.
The \textbf{binary reward} aligns with the coarse deployment decision:
\begin{equation}
  r_{\text{bin}}(\hat{s}, s) =
  \begin{cases}
    \phantom{-}\beta & \text{if } \mathbf{1}_{\text{same}}(\hat{s}, s) = 1, \\
    -\beta           & \text{otherwise},
  \end{cases}
\end{equation}
with $\beta > 0$.

\subsection{Optimization Objective}
\label{sec:grpo}

We train $\pi_\theta$ with GRPO~\cite{shao2024deepseekmath,schulman2017proximal}. For each training sample, we construct a prompt $\tilde{q}$ (with or without a hint prefix according to the schedule in Section~\ref{sec:schedule}) and sample $n$ responses $\{\mathbf{o}_j\}_{j=1}^{n}$ from the current policy.
The group-normalized advantage is
$\hat{A}_j = (r_j - \mu_r)/(\sigma_r + \varepsilon_{\text{stab}})$,
where $\mu_r$ and $\sigma_r$ are the within-group mean and standard deviation.
The training objective is:
\begin{multline}
  \mathcal{L}(\theta) = -\frac{1}{n}\sum_{j=1}^{n}
  \min\!\Bigl(
    \rho_j(\theta)\hat{A}_j,\;
    \mathrm{clip}\bigl(\rho_j(\theta),1{-}\varepsilon,1{+}\varepsilon\bigr)\hat{A}_j
  \Bigr) \\
  + \eta\,\mathrm{KL}\!\left(\pi_\theta \,\|\, \pi_{\text{ref}}\right),
  \label{eq:objective}
\end{multline}
where $\varepsilon$ is the PPO clipping coefficient and $\eta$ weights the KL penalty
against the frozen reference $\pi_{\text{ref}}$.
The importance-sampling ratio
$\rho_j(\theta) = \tilde{\pi}_\theta(\mathbf{o}_j\mid\tilde{q}_j) /
\tilde{\pi}_{\theta_{\text{old}}}(\mathbf{o}_j\mid\tilde{q}_j)$
is computed exclusively over generated (non-hint) tokens
via the masked policy $\tilde{\pi}$ from Section~\ref{sec:injection}.

\section{Experiments}
\label{sec:experiments}
\subsection{Experimental Setup}
\label{sec:setup}
\paragraph{Dataset.}
We construct Chinese-English social media translation datasets from Chinese social platform and translate them with several open- or close-source LLMs, covering internet slang, culturally-loaded expressions, homophonic ciphers,
and community-specific humor.
Human annotations follow a four-point rubric (0--3) jointly developed by
native English-speaking annotators and professional translators,
covering semantic accuracy, cultural appropriateness, and format compliance.
The dataset is split into a training set of 13,128 samples and a validation set of 1,444 samples, both balanced across the four score levels.
Translation hints for each training sample are generated offline by the auxiliary model $\mathcal{M}_h$ (Section~\ref{sec:hint}) conditioned on the source, translation, and human score.
\paragraph{Evaluation metrics.}
We report four metrics on the validation set:
(1)~\textbf{Binary Acc}: accuracy of the coarse binary decision
    ($\{0,1\}$ vs.\ $\{2,3\}$), reflecting practical deployment utility;
(2)~\textbf{Cohen's $\kappa$}: inter-rater agreement between model predictions
    and human judgments on the four-class scale;
(3)~\textbf{EM}: four-class exact match accuracy;
(4)~\textbf{Acc@$k$}: per-class accuracy on score-$k$ samples ($k\in\{0,\dots,3\}$), where Acc@0 covers the most culturally demanding cases.

\paragraph{Baselines.}
We compare CuRIL against four categories of baselines.
\textit{Traditional metrics}: CometKiwi, XCOMET, and BERTScore, as characterized
in Section~\ref{sec:preliminary}.
\textit{Open-source LLMs}: Qwen3-32B, Qwen3.5-27B, and Qwen3-235B,
evaluated zero-shot.
\textit{Closed-source LLMs}: GPT-4o-mini~\cite{achiam2023gpt}, DeepSeek-V4~\cite{xu2026deepseek}, GLM-5~\cite{zeng2026glm},
GPT-5.5~\cite{singh2025openai}, and Gemini-3.1-Pro~\cite{google-gemini-3}, evaluated few-shot.
\textit{Training baselines on the same base models}:
(a) Supervised fine-tuning (\textbf{SFT}) on the training set with cross-entropy loss;
(b) \textbf{Naive GRPO}, which applies standard GRPO without hint injection.

\subsection{Implementation Details}
\label{sec:implementation}

\paragraph{Base models.}
We train CuRIL on Qwen~\cite{qwen3,yang2024qwen2} model families:
Qwen2.5-7B-Instruct, Qwen3-4B, and Qwen3-8B.
All models are initialized from their publicly released instruction-tuned checkpoints.

\paragraph{Training framework.}
All RL experiments are conducted using the \textbf{verl} framework~\cite{sheng2024hybridflow},
which provides efficient rollout generation and policy gradient computation
for large language model training.

\paragraph{Hyperparameters.}
We use $n = 8$ rollouts per sample, hint injection rate $p_0 = 0.8$ decaying to $0$ over $T = 400$ steps, learning rate $1{\times}10^{-6}$ with cosine schedule, and batch size $B = 128$.
All models are trained for up to 600 steps on 8 $\times$ A100 GPUs.
Naive GRPO uses identical hyperparameters except $p_0 = 0$;
SFT is trained for 3 epochs with learning rate $5{\times}10^{-5}$.
All training experiments are run three times with independent random seeds;
we report the average across runs.
The full hyperparameter configuration is provided in the supplementary material.

\subsection{Main Results}
\label{sec:main_results}

\begin{table*}[t!]
\footnotesize
\begin{tabular*}{\textwidth}{@{\extracolsep{\fill}}llccccccc@{}}
\toprule
\multirow{2}{*}{\textbf{Category}} & \multirow{2}{*}{\textbf{Model}}
  & \multirow{2}{*}{\textbf{Bin.\ Acc}} & \multirow{2}{*}{\textbf{Cohen's $\kappa$}} & \multirow{2}{*}{\textbf{EM}}
  & \multicolumn{4}{c}{\textbf{Per-class Accuracy}} \\
\cmidrule(l){6-9}
  & & & & & \textbf{Acc@0} & \textbf{Acc@1} & \textbf{Acc@2} & \textbf{Acc@3} \\
\midrule
\multirow{4}{*}{Qwen2.5-7B}
  & Base                          & 56.69 & 0.134 & 27.51 & 15.24 & 29.09 & \textbf{54.17} & 11.63 \\
  & \quad $+$ SFT                 & 65.10 & 0.302 & 38.85 & \textbf{45.98} & 31.30 & 41.00 & 37.12 \\
  & \quad $+$ Naive GRPO          & 65.37 & 0.308 & 37.19 & 28.25 & \textbf{53.74} & 35.46 & 31.30 \\
  & \quad $+$ CuRIL (ours) & \textbf{66.14} & \textbf{0.323} & \textbf{40.44} & 42.38 & 49.31 & 25.21 & \textbf{44.88} \\
\midrule
\multirow{4}{*}{Qwen3-4B}
  & Base                          & 57.90 & 0.157 & 31.82 & 15.77 & 18.99 & 52.37 & \textbf{39.94} \\
  & \quad $+$ SFT                 & 62.19 & 0.244 & 37.60 & \textbf{35.73} & 44.88 & 41.55 & 28.25 \\
  & \quad $+$ Naive GRPO          & \textbf{67.84} & \textbf{0.357} & 36.59 & 15.28 & 52.63 & \textbf{56.51} & 21.88 \\
  & \quad $+$ CuRIL (ours) & 66.55 & 0.331 & \textbf{39.14} & 34.90 & \second{55.28} & 38.72 & 27.70 \\
\midrule
\multirow{4}{*}{Qwen3-8B}
  & Base                          & 57.75 & 0.147 & 31.60 & 18.12 & 11.57 & \second{59.20} & 36.69 \\
  & \quad $+$ SFT                 & 63.34 & 0.267 & 40.06 & 44.17 & 34.35 & 41.27 & \textbf{40.44} \\
  & \quad $+$ Naive GRPO          & 66.25 & 0.325 & 41.55 & 31.28 & \best{59.00} & 32.78 & 41.55 \\
  & \quad $+$ CuRIL (ours) & \second{69.11} & \textbf{0.370} & \second{45.22} & \best{54.01} & 54.29 & 41.82 & 36.84 \\
\midrule
\multirow{3}{*}{\shortstack[l]{Open-source\\LLMs}}
  & Qwen3-32B-Think               & 59.25 & 0.184 & 33.50 & 19.77 & 21.19 & 38.27 & 54.65 \\
  & Qwen3.5-27B-Think             & 68.70 & \second{0.373} & 42.02 & 39.76 & 25.71 & 29.55 & \best{72.42} \\
  & Qwen3-235B-A22B               & 58.10 & 0.162 & 31.72 & 16.07 & 24.93 & 33.52 & 52.35 \\

\midrule
\multirow{5}{*}{\shortstack[l]{Closed-source\\LLMs}}
  & GPT-4o-mini                   & 61.98 & 0.240 & 31.79 & 17.17 & 45.15 & 41.55 & 23.27 \\
  & DeepSeek-V4-Flash             & 64.22 & 0.285 & 39.46 & 36.01 & 21.05 & 31.30 & \second{69.64} \\
  & Gemini-3.1-Pro-Low            & \best{71.95} & \best{0.438} & \best{46.11} & 46.94 & 29.41 & 40.83 & 67.04 \\
  & GLM-5                         & 67.17 & 0.344 & 38.71 & \second{53.74} & 42.11 & 32.69 & 26.32 \\
  & GPT-5.5                       & 66.90 & 0.338 & 37.05 & 16.62 & 26.59 & \best{86.15} & 18.84 \\
\bottomrule
\end{tabular*}
\caption{Main results on the 1,444-sample social media translation validation set.
Bin.\ Acc: binary classification accuracy (\%);
$\kappa$: Cohen's kappa;
EM: four-class exact match accuracy (\%);
Acc@0--Acc@3: per-class accuracy (\%).
\textbf{Bold}: best trained variant per base model.
\protect\colorbox{aaibest}{\textbf{Green-shaded}}: best overall per column; \protect\colorbox{aaisecond}{\textbf{yellow-shaded}}: second-best.}
\label{tab:main}
\end{table*}

Table~\ref{tab:main} reports results across all models and baselines.

\paragraph{CuRIL consistently outperforms all training baselines.}
Across all three base model families,
CuRIL achieves the highest EM among trained models.
On Qwen3-8B, CuRIL reaches $\kappa = 0.370$ and EM $= 45.22\%$,
outperforming Naive GRPO by $+3.67\%$ in EM and $+0.045$ in $\kappa$.
Compared to SFT, CuRIL achieves higher EM ($+5.16\%$) and $\kappa$ ($+0.103$)
while requiring no labeled reasoning chains.
The advantage of CuRIL over Naive GRPO is most pronounced in EM,
indicating that cultural reasoning internalization primarily improves
the model's ability to make fine-grained four-class distinctions,
beyond the coarse binary classification that Naive GRPO already partially captures.

\paragraph{CuRIL matches frontier closed-source models at 8B scale.}
Our best model (Qwen3-8B + CuRIL) achieves $\kappa = 0.370$ and EM $= 45.22\%$,
approaching Gemini-3.1-Pro ($\kappa = 0.438$, EM $= 46.11\%$)
while using a model ${\sim}30\times$ smaller.
It substantially outperforms GPT-5.5 ($\kappa = 0.338$, EM $= 37.05\%$),
DeepSeek-V4-Flash ($\kappa = 0.285$, EM $= 39.46\%$),
and GLM-5 ($\kappa = 0.344$, EM $= 38.71\%$).
Notably, Qwen3-235B, a model with 235B total parameters, achieves
$\kappa = 0.162$ and EM $= 31.72\%$,
far below our 8B model,
confirming that model scale alone cannot compensate for the absence of
targeted cultural reasoning training.

\paragraph{Improvements are consistent across model families.}
CuRIL yields gains over Naive GRPO in EM for all three base models
(Qwen2.5-7B: $+3.25\%$; Qwen3-4B: $+2.55\%$; Qwen3-8B: $+3.67\%$),
and the relative benefit is largest for the strongest base model,
suggesting that cultural reasoning internalization scales favorably with
the model's underlying reasoning capacity.

\section{Analysis}
\label{sec:analysis}

\subsection{Alignment with Human Judgment}
\label{sec:analysis1}

We verify that Qwen3-8B + CuRIL achieves substantially stronger alignment
with human cultural judgment on the same validation set.

\paragraph{Correlation with human judgment.}
Table~\ref{tab:analysis_corr} compares correlation metrics across all evaluated methods.
Qwen3-8B + CuRIL achieves Pearson $r = 0.494$, Spearman $\rho = 0.492$,
and Cohen's $\kappa = 0.370$, respectively $2.3\times$ and $5.2\times$ higher than
the best traditional metric (CometKiwi) on Pearson $r$ and $\kappa$.

\paragraph{Sensitivity to error severity.}
Table~\ref{tab:analysis_severity} examines how each metric responds
as error severity increases.
Human scores decrease monotonically from 2.326 to 0.287 ($-$87.7\%),
while all three traditional metrics exhibit severity inversion:
XCOMET \emph{rises} from 0.568 to 0.743.
Qwen3-8B + CuRIL breaks this pattern: its scores fall monotonically from 1.704 to 0.726
($-$57.4\%), making it the only metric that correctly identifies severe errors as such.

\begin{table}[!ht]
\centering
\small
\begin{tabular*}{\columnwidth}{@{\extracolsep{\fill}}lccc@{}}
\toprule
\textbf{Metric} & \textbf{Pearson $r$} & \textbf{Spearman $\rho$} & \textbf{Cohen's $\kappa$} \\
\midrule
CometKiwi       &  0.212 &  0.222 &  0.071 \\
XCOMET          &  0.055 &  0.108 &  0.010 \\
BERTScore       & $-$0.115 & $-$0.118 & $-$0.003 \\
\midrule
\textbf{Qwen3-8B + CuRIL} & \textbf{0.494} & \textbf{0.492} & \textbf{0.370} \\
\bottomrule
\end{tabular*}
\caption{Correlation between metrics and human judgment.}
\label{tab:analysis_corr}
\end{table}

\begin{table}[!ht]
\centering
\small
\setlength{\tabcolsep}{4pt}
\begin{tabular*}{\columnwidth}{@{\extracolsep{\fill}}lccccc@{}}
\toprule
\textbf{Severity} & \textbf{Human} & \textbf{CometK.} & \textbf{XCOMET} & \textbf{BERT} & \textbf{CuRIL} \\
\midrule
No Error  & 2.326 & 0.532 & 0.568 & 0.573 & \textbf{1.704} \\
Minor     & 1.818 & 0.611 & 0.765 & 0.622 & \textbf{1.263} \\
Moderate  & 1.190 & 0.603 & 0.764 & 0.615 & \textbf{1.190} \\
Severe    & 0.287 & 0.569 & 0.743 & 0.622 & \textbf{0.726} \\
\midrule
Trend     & $\downarrow$ & $\uparrow$ & $\uparrow$ & $\uparrow$ & $\downarrow$ \\
\bottomrule
\end{tabular*}
\caption{Mean metric scores by human-rated error severity.}
\label{tab:analysis_severity}
\end{table}

\subsection{Hint Internalization Effect}
\label{sec:analysis2}

We validate that CuRIL enables models to internalize cultural reasoning,
making test-time hints unnecessary.
Three conditions are compared:
(1) \textbf{Base}: zero-shot without hints;
(2) \textbf{CuRIL}: trained model without hints;
(3) \textbf{+tp}: base model with hints at test time (idealized upper bound).

\begin{figure}[!ht]
    \centering
    \includegraphics[width=\columnwidth]{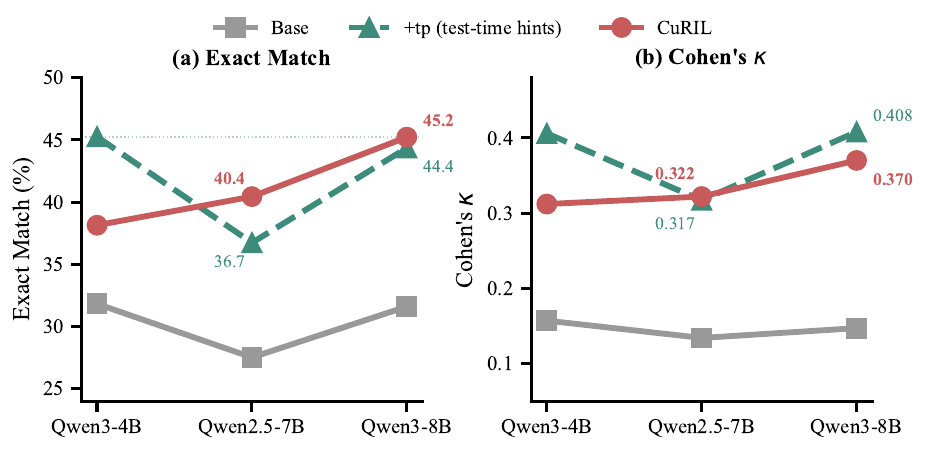}
    \caption{Hint internalization across model scales. CuRIL (solid red) surpasses test-time hint injection (+tp, dashed teal) at 7B+ scale, demonstrating genuine internalization of cultural reasoning.}
    \label{fig:internalization}
\end{figure}

As shown in Figure~\ref{fig:internalization}, a scale-dependent pattern emerges.
At 4B scale, test-time hints still outperform internalization
(EM 45.25\% vs.\ 38.14\%), suggesting insufficient capacity.
However, at 7B+ scale the picture reverses:
Qwen2.5-7B after CuRIL achieves EM 40.44\%, surpassing hints (36.73\%),
and Qwen3-8B after internalization (EM 45.22\%) surpasses the hint-augmented
base model (44.35\%) while requiring no external input at inference time.

\subsection{CuRIL as a Reward Model}
\label{sec:analysis3}

\begin{figure}[!ht]
    \centering
    \includegraphics[width=\columnwidth]{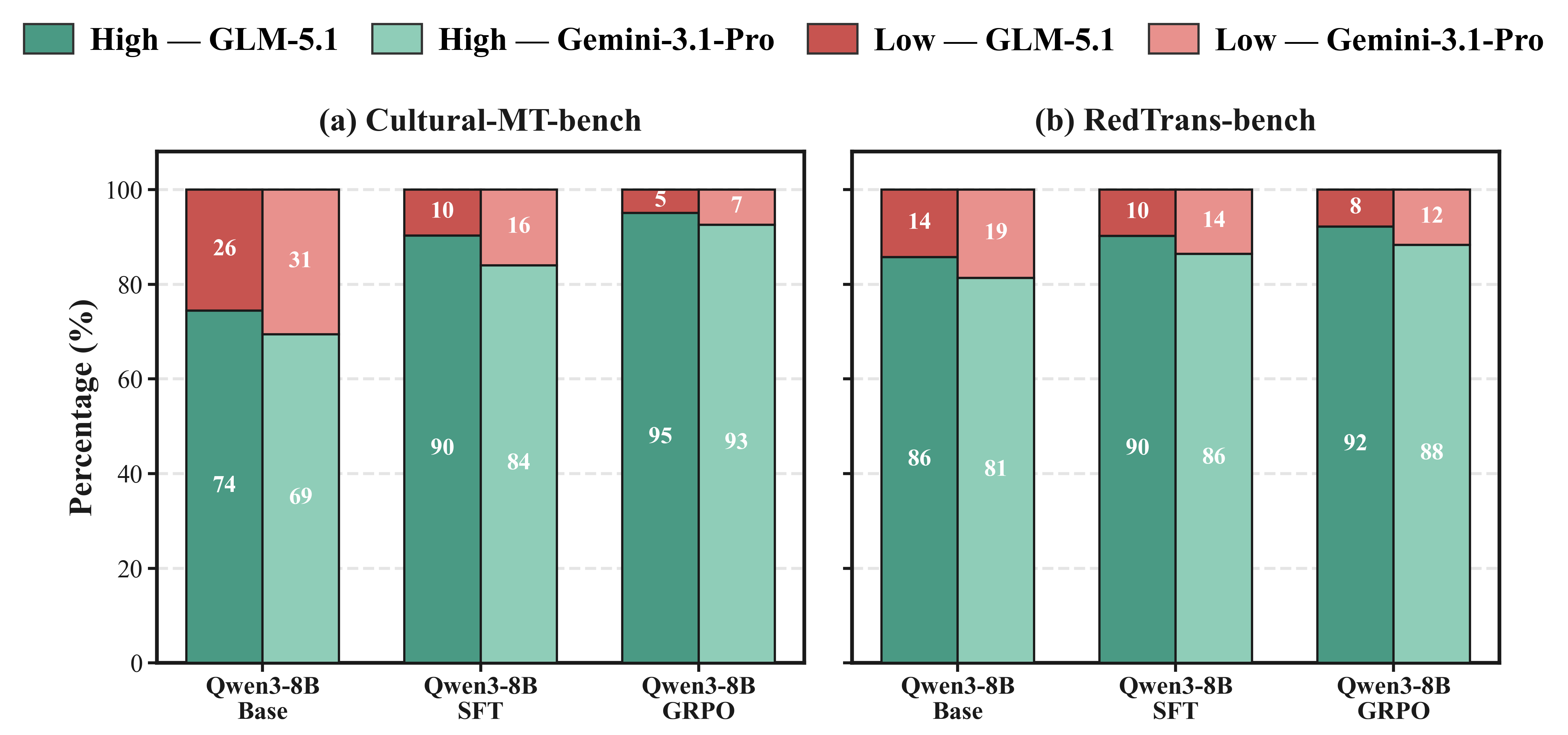}
    \caption{Low- and high-quality proportions of Base, SFT, and GRPO on Cultural-MT Bench and RedTrans-Bench, judged by GLM-5 (solid) and Gemini-3.1-Pro (hatched).}
    \label{fig:downstream}
\end{figure}

Beyond evaluation, we validate the CuRIL judge as a reward signal
for training downstream translation models via RLVR.
The training data for the downstream translator is constructed following Cultural-MT Bench~\cite{wu2026beyond}.
We evaluate Qwen3-8B as a translation model under three conditions (Base, SFT, and GRPO with CuRIL reward) on two held-out benchmarks:
Cultural-MT Bench (1,002 social note samples)
and RedTrans-Bench (2,858 short note or comment samples),
judged by two independent evaluators, GLM-5 and Gemini-3.1-Pro.
As shown in Figure~\ref{fig:downstream},
GRPO with CuRIL reward consistently outperforms both Base and SFT.
On Cultural-MT Bench, the low-quality rate drops from 25.6\% (Base) to 4.9\% (GLM-5) and from 30.6\% to 7.4\% (Gemini-3.1-Pro), both well below SFT (9.7\% and 16.0\%).
The improvement generalizes to RedTrans-Bench, confirming that CuRIL provides a reliable reward signal beyond SFT alone.

\subsection{Training Dynamics}
\label{sec:analysis4}

\begin{figure}[!ht]
    \centering
    \includegraphics[width=\columnwidth]{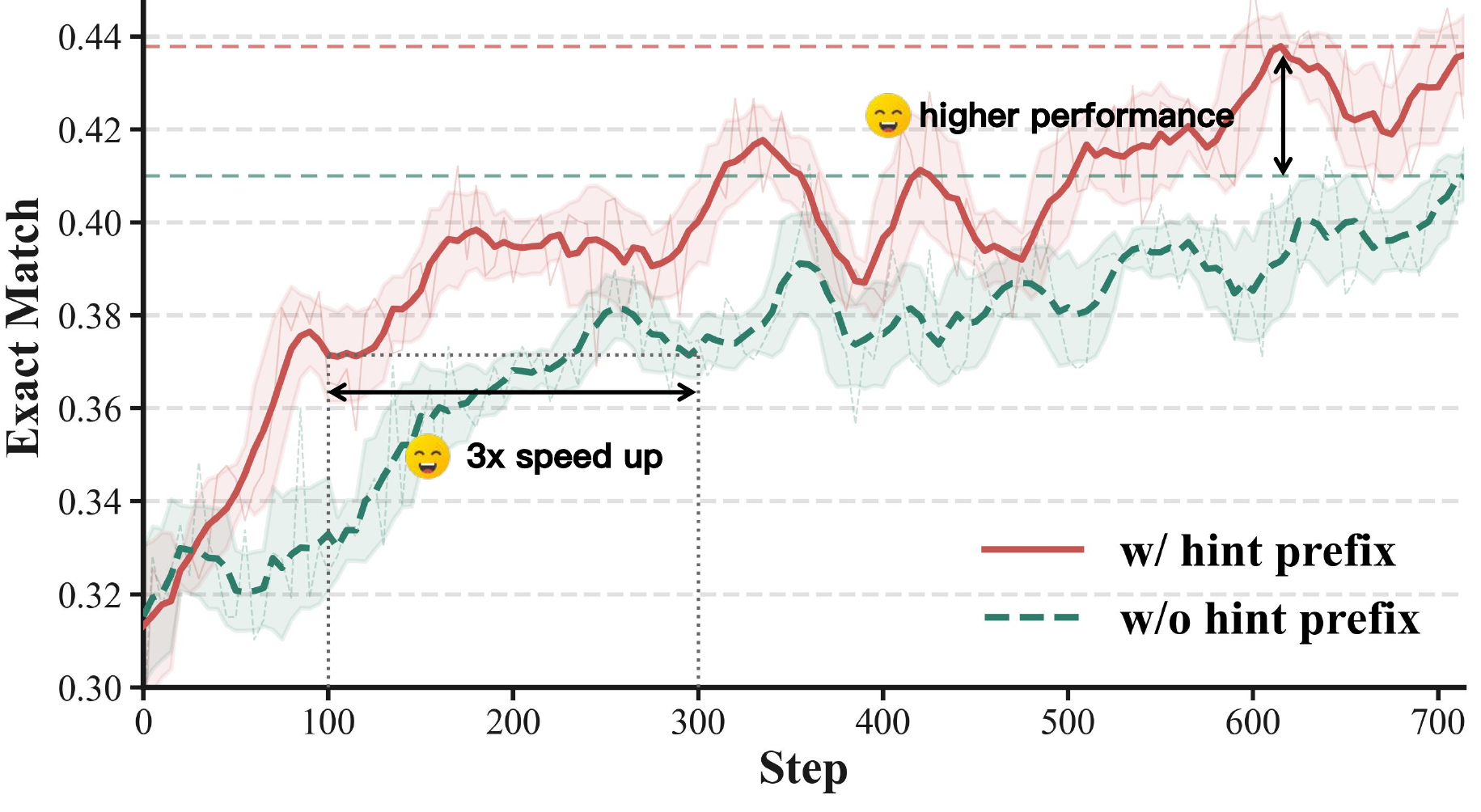}
    \caption{Validation EM of CuRIL (solid) and Naive GRPO (dashed) on Qwen3-8B over training; shaded regions show local std.}
    \label{fig:training_dynamics}
\end{figure}

Figure~\ref{fig:training_dynamics} compares the validation EM curves
of CuRIL and Naive GRPO on Qwen3-8B.

\paragraph{Faster convergence.}
CuRIL surpasses 37.5\% EM within approximately 100 steps,
while Naive GRPO requires nearly 300 steps, a $3\times$ reduction in steps to threshold.

\paragraph{Higher performance ceiling.}
Naive GRPO plateaus at approximately 41\% EM;
CuRIL continues past this barrier, reaching roughly 45\%,
suggesting that cultural reasoning signals not only accelerate training but expand the strategies accessible to the policy.

\subsection{Cross-Domain Generalization}
\label{sec:analysis5}

To assess generalizability beyond our training domain, we evaluate on the
MENT dataset~\cite{tian2026ment}, a meta-evaluation benchmark for non-literal
translation covering SNS, cross-culture, poetry, and literature domains.
Because MENT uses a 5-point scale while our models are trained on a
4-point scale, we report only rank-correlation coefficients in Table~\ref{tab:ment}.

\begin{table}[!ht]
\centering
\small
\begin{tabular*}{\columnwidth}{@{\extracolsep{\fill}}lccc@{}}
\toprule
\textbf{Model} & \textbf{Pearson $r$} & \textbf{Spearman $\rho$} & \textbf{Kendall $\tau$} \\
\midrule
Qwen2.5-7B-Instruct        & 0.51 & 0.51 & 0.42 \\
\quad $+$ Naive GRPO       & 0.53 & 0.52 & 0.43 \\
\quad $+$ CuRIL            & \textbf{0.56} & \textbf{0.56} & \textbf{0.46} \\
\midrule
Qwen3-4B                   & 0.45 & 0.46 & 0.37 \\
\quad $+$ Naive GRPO       & 0.46 & 0.47 & 0.39 \\
\quad $+$ CuRIL            & \textbf{0.48} & \textbf{0.48} & \textbf{0.39} \\
\bottomrule
\end{tabular*}
\caption{Correlation with human judgments on the MENT dataset.}
\label{tab:ment}
\end{table}

CuRIL consistently improves correlation with human judgments over both the base model and Naive GRPO across all metrics,
demonstrating that cultural reasoning internalized on Chinese--English social media transfers to broader non-literal translation evaluation settings.

\section{Related Work}
\label{sec:related}

\subsection{Translation Quality Evaluation}
\label{sec:related1}

Automatic metrics have evolved from n-gram overlap~\cite{papineni2002bleu} to learned similarity: BERTScore~\cite{zhang2019bertscore} uses contextual embeddings, and the COMET family~\cite{rei2020comet,rei2022cometkiwi,guerreiro2023xcomet} learns from human judgments.
All share one inductive bias, source--translation alignment as a proxy for quality, which breaks down on social media where literal renderings of culturally loaded expressions achieve high overlap yet convey no meaning.
LLM judges offer richer evaluation: GEMBA~\cite{kocmi2023gemba} reaches near human-level agreement on general-domain MT; MENT~\cite{tian2026ment} shows that both metrics and static judges fail on non-literal domains and compensates by retrieving knowledge at inference; CULTURE-MT~\cite{wu2026beyond} trains an SFT judge for cultural effectiveness on Chinese social media UGC.
Cultural-competence studies further show that scale alone does not close cultural gaps~\cite{liu2026gaoyao,yao2024benchmarking}, and RedTrans~\cite{guo2025redefining} targets SNS translation with a 72B domain-adapted model, whose RedTrans-Bench we adopt for out-of-domain evaluation.
We share these diagnoses but differ in the remedy: rather than retrieving at inference~\cite{tian2026ment,wang2024retrieval,conia2024towards,agrawal2023context} or relying on annotated SFT alone~\cite{wu2026beyond}, CuRIL internalizes cultural reasoning into the judge via reinforcement learning, requiring no external input at deployment.

\subsection{Hint-Guided Reinforcement Learning}
\label{sec:related3}

RLHF~\cite{ouyang2022training,lambert2024tulu} established preference-aligned policy optimization; GRPO~\cite{shao2024deepseekmath} simplifies it with group-normalized advantages, and DeepSeek-R1~\cite{guo2025deepseek} shows that verifiable rewards elicit strong reasoning.
A recent line of work guides RL exploration with hints injected during training~\cite{su2025trust,wang2026milestone}: StepHint~\cite{zhang2026stephint} provides stepwise hints for mathematical reasoning, HintGRPO~\cite{huang2025boosting} applies debiased hints to multimodal and agentic tasks~\cite{ALFWorld20,Yao2022WebShopTS,boiko2023emergent}, RuscaRL~\cite{zhou2026breakingexplorationbottleneckrubricscaffolded} places checklist rubrics in the task instruction with decaying strength, and Scaf-GRPO~\cite{zhang2026scafgrposcaffoldedgrouprelative} injects tiered in-prompt hints when learning plateaus.
All deliver guidance on the \emph{input side} for tasks with verifiable intermediate structure.
CuRIL differs in both bottleneck and mechanism: what is missing here is domain-specific cultural knowledge unavailable at inference, so hints are injected \emph{inside the model's own reasoning} as a gradient-masked response prefix and decayed to zero, forcing the knowledge itself to be internalized.

\section{Conclusion}
\label{sec:conclusion}

We showed that surface-similarity metrics systematically reward the worst social media translations, and proposed CuRIL, which internalizes cultural reasoning by injecting translation hints as a gradient-masked, linearly decaying prefix during reinforcement learning.
CuRIL lifts Qwen3-8B to $\kappa = 0.370$ and 45.22\% EM, approaching Gemini-3.1-Pro, and its reward signal cuts the downstream low-quality translation rate from 25.6\% to 4.9\%.
More broadly, domain-specific evaluation ability need not be supplied at inference: it can be internalized through decaying guidance, a recipe applicable beyond translation.

\bibliography{aaai2027}

\clearpage

\appendix
\setcounter{figure}{0}
\setcounter{table}{0}
\renewcommand{\thefigure}{S\arabic{figure}}
\renewcommand{\thetable}{S\arabic{table}}

\section{Ablation Study on Hint Decay Schedule}
\label{app:ablation_decay}

We ablate the hint injection schedule to isolate the contribution of the linear decay
curriculum.
Three variants of CuRIL are compared on Qwen3-8B:

\begin{itemize}
  \item \textbf{CuRIL} (linear decay): hint injection probability decays linearly
        from $p_0 = 0.8$ to $0$ over $T = 400$ steps, then reduces to standard GRPO.
  \item \textbf{Without decay}: hint injection probability is held constant at $p_0 = 0.8$
        throughout training. Note that this variant collapses after approximately 400 steps,
        suggesting that sustained hint availability prevents the model from developing
        autonomous cultural reasoning and eventually destabilizes the reward landscape.
  \item \textbf{Self-hint}: no offline hints are provided; instead, the model is prompted
        to generate its own cultural key points before scoring, testing whether
        self-generated reasoning signals can substitute for curated hints.
\end{itemize}

\begin{table}[h]
\centering
\small
\begin{tabular*}{\columnwidth}{@{\extracolsep{\fill}}lcccc@{}}
\toprule
\textbf{Variant} & \textbf{Bin.\ Acc} & $\boldsymbol{\Delta}$ & \textbf{EM} & $\boldsymbol{\Delta}$ \\
\midrule
CuRIL (linear decay)       & \best{69.11} & ---     & \best{45.22} & ---     \\
Without decay              & 65.81          & $-$3.30 & 40.29          & $-$4.93 \\
Self-hint                  & \second{67.52} & $-$1.59 & \second{41.41} & $-$3.81 \\
\bottomrule
\end{tabular*}
\caption{Ablation on hint injection schedule (Qwen3-8B). Linear decay consistently
         outperforms both alternatives. Without-decay collapses after ${\sim}400$ steps.}
\label{tab:ablation_decay}
\end{table}

Both alternatives underperform the linear decay schedule.
Without decay, the model never faces hint-free rollouts during the decay phase
and thus fails to internalize cultural reasoning---collapsing once the held-out
evaluation reveals its hint dependence.
Self-hint performs better than without-decay but still lags behind CuRIL,
indicating that model-generated hints lack the precision and grounding
of offline curated annotations, even though self-generated reasoning
provides some directional benefit.

\section{Ablation Study on Gradient Masking}
\label{app:ablation_mask}

A core design choice in CuRIL is the gradient mask applied to the hint prefix:
policy gradients are computed only over the model-generated tokens, while the
prepended hint tokens are excluded from the loss.
To validate this design, we compare CuRIL against a variant that removes the
gradient mask, allowing the policy gradient to flow through the entire
concatenated sequence---including the externally generated hint prefix.

\begin{figure}[h]
\centering
\includegraphics[width=\columnwidth]{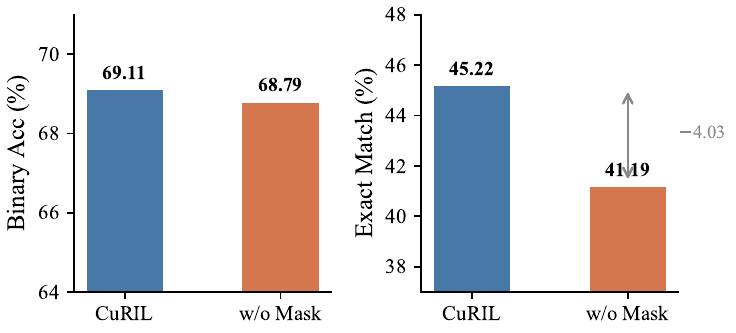}
\caption{Ablation on gradient masking (Qwen3-8B). Removing the mask causes a
         substantial drop in EM while Binary Accuracy remains comparable.}
\label{fig:ablation_mask}
\end{figure}

Removing the gradient mask leads to a notable 4.03-point decline in Exact Match
accuracy while Binary Accuracy decreases only marginally ($-$0.32).
Without the mask, the model is trained to reproduce the externally generated hint
tokens as part of its own output, conflating two distinct objectives: faithfully
copying an external annotation and learning to reason about cultural content
autonomously.
This forces the policy to allocate capacity toward imitating hint text that will
be absent at inference time, degrading its ability to make fine-grained four-class
distinctions---hence the disproportionate drop in EM relative to the coarser
binary metric.
The result confirms that gradient masking is essential for CuRIL to treat hints
as exploration guidance rather than supervision targets.

\section{Ablation Study on Decay Function}
\label{app:ablation_decay_func}

CuRIL uses a linear schedule to decay the hint injection probability from
$p_0 = 0.8$ to $0$ over $T$ training steps.
We compare this default against two alternatives:
cosine decay ($p(t) = \tfrac{p_0}{2}(1 + \cos(\pi t / T))$) and
exponential decay ($p(t) = p_0 \cdot e^{-\lambda t}$, calibrated so that
$p(T) \approx 0.01$).

\begin{figure}[h]
\centering
\includegraphics[width=\columnwidth]{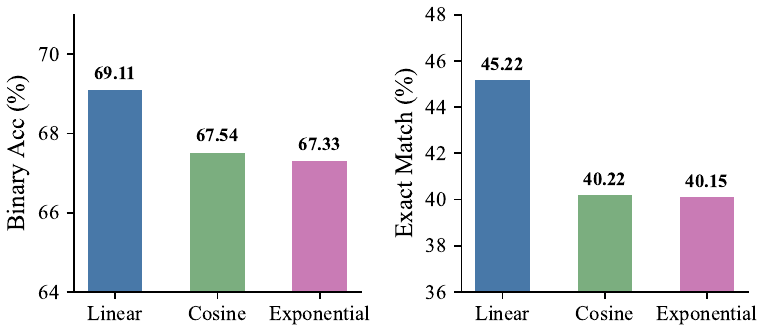}
\caption{Ablation on decay function (Qwen3-8B). Linear decay outperforms both
         cosine and exponential alternatives.}
\label{fig:ablation_decay_func}
\end{figure}

Both non-linear schedules underperform the linear baseline by a substantial
margin (${\sim}5$ EM points).
Cosine and exponential decay reduce the injection probability slowly at first
and then drop sharply near the end of the schedule, meaning the model receives
hints at a high rate for the majority of training and then faces an abrupt
transition to hint-free rollouts.
Linear decay, by contrast, steadily increases the fraction of unassisted rollouts
from the outset, providing a more uniform curriculum that gives the model
consistent practice at autonomous cultural reasoning throughout training.
The simplest schedule thus proves the most effective.

\section{Translation Hint Extraction}
\label{app:hint_extraction}

As described in the main paper, translation hints are generated offline
by an auxiliary LLM $\mathcal{M}_h$.
Concretely, for each training sample we provide $\mathcal{M}_h$ with the
source comment, machine translation, surrounding post context,
and the human reviewer's assessment (including error spans, severity labels,
and corrective remarks).
$\mathcal{M}_h$ is instructed to distill concise translation key points
from the reviewer's assessment---each point 10--30 characters,
covering only issues genuinely present in the review.
If the translation is error-free, the output is simply ``None.''
The prompt template used for hint extraction is shown below.

\begin{tcolorbox}[
    title=Prompt Template for Translation Hint Extraction (translated from Chinese),
    breakable
]
\small
\textbf{System:} You are a senior Chinese--English translation expert skilled at distilling key translation points from review comments.\\[4pt]
\textbf{User:} I have a translation quality assessment entry. Please extract concise translation key points from the review comments.\\[4pt]
Requirements:\\
1. Each point should be 10--30 characters; list them as \textbf{[Point $N$]}.\\
2. Only list issues that genuinely appear in the review; do not fabricate.\\
3. If the translation is perfect, output ``Translation points: None.''\\
4. Output only the translation points section; no other content.\\[4pt]
Output format:\\
Translation points:\\
{[Point 1]} xxx\\
{[Point 2]} xxx\\
...\\[4pt]
---\\
{[Data below]}\\[2pt]
Context: \{context\}\\
Source: \{content\}\\
Translation: \{trans\_result\}\\
Review comments: \{original\_value\}\\
---\\[2pt]
Please output the translation points directly (no preamble):
\end{tcolorbox}

The extraction is parallelized across training samples with checkpoint-based
resumption.
Each resulting hint string is stored in the \texttt{translation\_points} field
of the training data and used exclusively as the gradient-masked prefix
during CuRIL training.

\section{Data Format}
\label{app:data_format}

Each training sample consists of the following fields:

\begin{itemize}
  \item \textbf{content}: the original Chinese social media comment, wrapped in
        XML-style segment tags (e.g., \texttt{<comment3>}).
  \item \textbf{trans\_result}: the machine-translated English output, wrapped in
        the corresponding tag.
  \item \textbf{score}: the human quality score ($0$--$3$).
  \item \textbf{context}: surrounding post context, including the original post
        and hashtags, providing pragmatic grounding for the comment.
  \item \textbf{error\_list}: a list of annotated errors, each specifying the
        erroneous text span, severity level, error category, and a corrective remark.
  \item \textbf{translation\_points}: offline-generated cultural hints summarizing
        the key translation challenges for the sample, used as hint prefix during
        CuRIL training.
\end{itemize}

A representative example is shown below (Chinese text romanized for typesetting).
The idiom \textit{jin bu huan} (``priceless / worth a thousand gold'')
is mistranslated literally as ``Gold never changes'' (score~0).
The \texttt{translation\_points} field provides the cultural hint prefix
injected during CuRIL training.

\begin{tcolorbox}[
    title=Example Training Sample,
    breakable
]
\small
\{\\
\quad ``content'': ``$<$comment3$>$[Jin bu huan haha]$<$/comment3$>$'',\\
\quad ``trans\_result'': ``$<$comment3$>$Gold never changes, haha$<$/comment3$>$'',\\
\quad ``score'': 0,\\
\quad ``context'': ``0.[Are all dogs at the beach this excited?]\\
\qquad 1.[I can feel the joy of the little dog]\\
\qquad \#[Dogs by the sea][topic]\# ...'',\\
\quad ``error\_list'': [\\
\qquad \{\\
\qquad\quad ``error\_text'': ``Gold never changes'',\\
\qquad\quad ``error\_degree'': ``severe'',\\
\qquad\quad ``error\_tag'': ``semantic deviation'',\\
\qquad\quad ``error\_remark'': ``[Jin bu huan] is a Chinese idiom meaning something priceless and irreplaceable.''\\
\qquad \}\\
\quad ],\\
\quad ``translation\_points'': ``[Translation hints:\\
\qquad [Point 1] The idiom `jin bu huan' denotes priceless value; literal translation is not acceptable.\\
\qquad [Point 2] Convey the nuance of `irreplaceable' fitting the context.]''\\
\}
\end{tcolorbox}

\section{Judger System Prompt}
\label{app:prompt}

The following system prompt is used for all judger model evaluations,
including baseline zero-shot LLMs and our trained CuRIL models.

\begin{tcolorbox}[
    title=Judger System Prompt (translated from Chinese),
    breakable
]
\small
You are a bilingual (Chinese--English) expert translation quality evaluator with deep familiarity with internet culture, applying exceptionally strict standards. Your task is to evaluate a translation given its context, source, and target text according to the following rules.\\[4pt]
1. Scoring criteria:\\
\quad Semantic accuracy:\\
\quad - Correctly interpret source semantics and emotion (irony, excitement, etc.)\\
\quad - Culturally appropriate address terms (e.g., ``jiemei'', ``laoshi'', ``baozi'')\\
\quad - Target-culture naturalness; avoid Chinglish\\
\quad - Culture-loaded terms (e.g., ``juejuezi'', ``zhongcao''): must be sense-translated, not literally\\
\quad - Unit conversions (e.g., mu, jin, li): accurate and unambiguous\\
\quad - Proper nouns: use accurate or established translations\\[4pt]
\quad Format rules (violation = immediate 0):\\
\quad - Placeholder tokens (\#zhanwei-x\#) must NOT be translated\\
\quad - XML tags ($<$comment$>$, $<$title$>$, $<$content$>$, etc.) must be preserved and matched\\[4pt]
2. Score definitions:\\
\quad 0 -- Severe error: source meaning lost or fundamentally distorted\\
\quad 1 -- Noticeable problems: main meaning barely recoverable; key errors present\\
\quad 2 -- Adequate: main information conveyed; minor unnatural expression or cultural mismatch\\
\quad 3 -- Excellent: accurate, natural, culturally appropriate, publication-ready\\[4pt]
3. Output format:\\
\quad $<$think$>$reasoning process$<$/think$>$\\
\quad $<$answer$>$score$<$/answer$>$\\[4pt]
Evaluate the following translation strictly according to the above guidelines. Output only the evaluation; no additional text.
\end{tcolorbox}

\section{Baseline Implementation Details}
\label{app:baselines}

To ensure a fair comparison, all training-based baselines are matched to CuRIL in terms of data and compute budget.

\paragraph{Supervised Fine-Tuning (SFT).}
SFT is trained on the same 13,128-sample training set used for CuRIL and Naive GRPO.
To maintain a controlled comparison, the training is run for the same number of steps as the RL-based methods (600 steps), with a learning rate of $5\times10^{-5}$ and a cosine schedule.
The model is trained with standard cross-entropy loss on (query, score) pairs,
where the query consists of the source--translation pair formatted with the same system prompt used during RL training.
No hints or auxiliary annotations are provided during SFT.

\paragraph{Naive GRPO.}
Naive GRPO applies standard Group Relative Policy Optimization~\cite{shao2024deepseekmath} to the same training data and for the same number of steps (600) as CuRIL.
All hyperparameters---including rollout count ($n=8$), learning rate, PPO clipping coefficient, and reward function---are identical to CuRIL.
The sole difference is that the hint injection probability is fixed at $p_0 = 0$ throughout training: no cultural hints are prepended to any rollout at any step.
Naive GRPO therefore serves as a direct ablation of the hint injection mechanism, isolating the contribution of cultural reasoning signals from other design choices in CuRIL.

\newpage
\section{Hyperparameter Configuration}
\label{app:hyperparams}

Table~\ref{tab:hyperparams} lists the full hyperparameter configuration
used for all CuRIL training runs.

\begin{table}[H]
\centering
\footnotesize
\renewcommand{\arraystretch}{1.15}
\begin{tabular}{@{}lll@{}}
\toprule
\textbf{Category} & \textbf{Parameter} & \textbf{Value} \\
\midrule
\multirow{5}{*}{Data}
  & \texttt{train\_batch\_size}              & 128 \\
  & \texttt{max\_prompt\_len}               & 2048 \\
  & \texttt{max\_response\_len}             & 4096 \\
  & \texttt{filter\_overlong\_prompts}      & True \\
  & \texttt{truncation}                     & \texttt{error} \\
\midrule
\multirow{8}{*}{\makecell[l]{Model /\\Actor}}
  & \texttt{optim.lr}                       & $1\times10^{-6}$ \\
  & \texttt{ppo\_mini\_batch\_size}         & 32 \\
  & \texttt{ppo\_micro\_batch\_size}        & 2 \\
  & \texttt{use\_kl\_loss}                  & True \\
  & \texttt{kl\_loss\_coef}                 & 0.001 \\
  & \texttt{kl\_loss\_type}                 & \texttt{low\_var\_kl} \\
  & \texttt{entropy\_coeff}                 & 0 \\
  & \texttt{grad\_checkpointing}            & True \\
\midrule
\multirow{5}{*}{Rollout}
  & \texttt{log\_prob\_micro\_batch\_size}   & 4 \\
  & \texttt{tensor\_parallel\_size}         & 4 \\
  & \texttt{name}                           & \texttt{vllm} \\
  & \texttt{gpu\_memory\_util}              & 0.4 \\
  & $n$ (rollouts per prompt)               & 8 \\
\midrule
\multirow{2}{*}{\makecell[l]{Hint\\Schedule}}
  & \texttt{hint.initial\_ratio}            & 0.8 \\
  & \texttt{hint.decay\_steps}              & 400 \\
\bottomrule
\end{tabular}
\caption{Full hyperparameter configuration for CuRIL training.}
\label{tab:hyperparams}
\end{table}

\end{document}